\documentclass[10pt,twocolumn,letterpaper]{article}

\usepackage{iccv}
\usepackage{times}
\usepackage{epsfig}
\usepackage{graphicx}
\usepackage{amsmath}
\usepackage{amssymb}

\usepackage{amsthm}
\usepackage{booktabs}
\usepackage[ruled,linesnumbered]{algorithm2e}

\usepackage[pagebackref=true,breaklinks=true,letterpaper=true,colorlinks,bookmarks=false]{hyperref}

\iccvfinalcopy 

\def\iccvPaperID{****} 

\ificcvfinal\fi

\begin{document}

\title{Group-CAM: Group Score-Weighted Visual Explanations for Deep Convolutional Networks}

\author{Qinglong Zhang, Lu Rao, Yubin Yang\\
	State Key Laboratory for Novel Software Technology at Nanjing University\\
	{\tt\small \{wofmanaf, raoluSmile\}@smail.nju.edu.cn, yangyubin@nju.edu.cn}
}


\maketitle
\ificcvfinal\thispagestyle{empty}\fi

\begin{abstract}
Recently, explaining deep convolutional neural networks has been drawing increasing attention since it helps to understand the networks' internal mechanisms and the reason why the networks make specific decisions. In the computer vision community, one of the most popular approaches for visualizing and understanding deep networks is generating saliency maps that highlight salient regions mostly related to the network's decision-making. However, saliency maps generated by existing methods either contain too much meaningless information or the process of computing saliency maps requires plenty of time. In this paper, we propose an efficient saliency map generation method, called Group score-weighted Class Activation Mapping (Group-CAM), which adopts the ``split-transform-merge" strategy to generate saliency maps. Specifically, for an input image, the class activations are firstly split into groups. In each group, the sub-activations are summed and de-noised as an initial mask. After that, the initial masks are transformed with meaningful perturbations and then applied to preserve sub-pixels of the input (i.e., masked inputs), which are then fed into the network to calculate the confidence scores. Finally, the initial masks are weighted summed to form the final saliency map, where the weights are confidence scores produced by the masked inputs. Group-CAM is efficient yet effective, which only requires dozens of queries to the network while producing target-related saliency maps. As a result, Group-CAM can be served as an effective data augment trick for fine-tuning the networks. We comprehensively evaluate the performance of Group-CAM on common-used benchmarks, including deletion and insertion tests on ImageNet-1k, and pointing game tests on COCO2017. Extensive experimental results demonstrate that Group-CAM achieves better visual performance than the current state-of-the-art explanation approaches. The code is available at https://github.com/wofmanaf/Group-CAM.

\end{abstract}

\section{Introduction}
\label{sec:intro}
Understanding and interpreting the decision made by deep neural networks (DNNs) is of central importance for humans since it helps to construct the trust of DNN models \cite{DBLP:conf/iccv/KapishnikovBVT19,DBLP:conf/cvpr/BansalAN20a,DBLP:conf/cvpr/RebuffiFJV20,DBLP:conf/cvpr/XuVS20}. 
In the area of computer vision, one critical technique is generating intuitive heatmaps that highlight regions, which are most related to DNN's decision.

One common approach for determining salient regions is relying on the changes in the model output, such as the changes of prediction scores with respect to the input images. For example, RISE~\cite{DBLP:conf/bmvc/PetsiukDS18} estimates importance empirically by probing the model with randomly masked versions of the image and obtaining the corresponding outputs. While RISE provides very compelling results, thousands of random masks should be generated and then be applied to query the model, making it inefficient.

Other approaches, such as Grad-CAM~\cite{DBLP:conf/iccv/SelvarajuCDVPB17}, calculate gradients by back-propagating the prediction score through the target layer of the network and apply them as weights to combine the forward feature maps. These methods are generally faster than RISE since they only require a single or constant number of queries to the network~\cite{DBLP:conf/aaai/QiKL20,DBLP:conf/iccv/KapishnikovBVT19}. However, saliency maps of Grad-CAM may capture too much meaningless information since the feature maps are not necessarily related to the target category~\cite{DBLP:conf/aaai/ZhangRY21}. Therefore, the results of
Grad-CAM may not truly reflect the reason why a network makes a decision. Naturally, a question arises: ``\textit{Can one method produce results that truly reflect the model decision in a more efficient way}?"

\begin{figure*}[htb]
	\centering
	\includegraphics[width=0.85\linewidth]{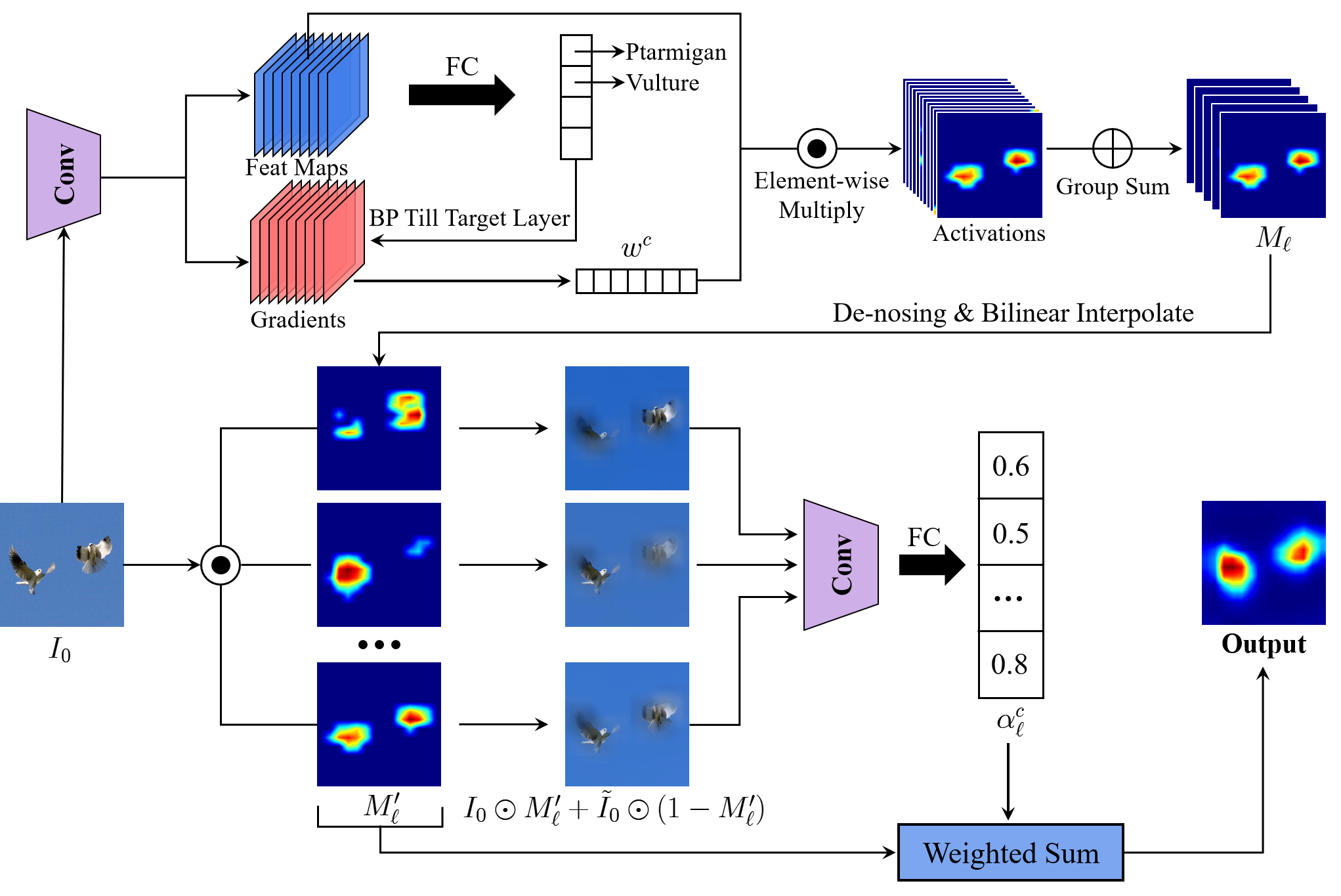}
	\caption{Pipeline of Group-CAM. Activations are first extracted with a linear combination of feature maps and importance weights $w^{c}$. Then the activations are split into groups and summed along the channel dimension in each group before de-noising to generate initial masks $M'_{\ell}$. Input image $I_0$ is element-wise multiplied with $M'_{\ell}$ and then transformed with meaningful perturbations. The perturbated images are then fed to the network. The output saliency map can be computed as a weighted sum of all $M'_{\ell}$ where the weights $\alpha_{\ell}^{c}$ come from the confidence scores of the target class corresponding to the respective perturbated inputs.}
	\label{fig:fig1}
\end{figure*}

To answer this question, we first revisit the intuition behind RISE~\cite{DBLP:conf/bmvc/PetsiukDS18}. Let $M$ be a random binary mask with distribution $\mathcal{D}$, the input image $I_0$ can be masked by $I_0 \odot M$ to preserve a subset of pixels, where $\odot$ denotes element-wise multiplication. The masked image is then applied to produce the confidence score to measure the contribution of these preserved pixels. Finally, the saliency map can be generated by combining plenty of random masks and scores with respect to them. It is observed that the most time-costing procedure is random masks generating and multiple queries to the neural network.

To address the efficiency issue, we propose Group score-weighted Class Activation Mapping (Group-CAM), which adopts the ``split-transform-merge" strategy to generate saliency maps. Specifically, for an input image, the class activations are firstly split into groups (to improve the efficiency, we apply the simplest uniform and adjacent group strategy, same as ResNexT~\cite{DBLP:conf/cvpr/XieGDTH17}). In each group, the sub-activations are summed along the channel dimension as an initial mask. However, directly apply the initial masks to preserve input pixels may cause noise visual due to gradient vanishing~\cite{DBLP:conf/aaai/ZhangRY21}. Therefore, we design a de-noising strategy to filter the less important pixels of the initial mask. In addition, to ease the adversarial effects of sharp boundaries between the masked and salient regions, we employ the blurred information from the input to replace the unreserved regions(pixels with 0 values) of the masked image. Finally, the saliency map of Group-CAM can be computed as a weighted sum of the grouped initial masks, where the weights are confidence scores produced by the masked inputs. Group-CAM is quite efficient, which can produce appealing target-related saliency maps after dozens of queries to the networks. As a result, Group-CAM can be applied to train/fine-tune classification methods. 
The overall architecture of Group-CAM are illustrated in Figure~\ref{fig:fig1}.


We comprehensively evaluate Group-CAM on ImageNet-1k and MS COCO2017. Results demonstrate that Group-CAM requires less computation yet achieves better visual performance than the current state-of-the-art methods. In addition, we extend the application of saliency methods and apply Group-CAM as an effective data augment trick for fine-tuning classification networks, extensive experimental results suggest that Group-CAM can boost the networks' performance by a large margin.

Note that, if the number of groups in Group-CAM is set to 1, and no de-noising strategy is applied, then Group-CAM can be simplified as Grad-CAM.

\section{Related Work}
\label{sec:ret_work}
\textbf{Region-based Saliency Methods.}
In recent years, numerous saliency methods attributing inputs to output predictions have been proposed~\cite{DBLP:conf/aaai/ZhangRY21,DBLP:conf/iccv/SelvarajuCDVPB17,DBLP:conf/iccv/KapishnikovBVT19}. One set of methods adopt masks to preserve certain regions of the inputs and measure the effect these regions have on the output by performing a forward pass through the network with these regions. These types of saliency methods are called Region-based saliency methods. Among them, RISE~\cite{DBLP:conf/bmvc/PetsiukDS18} first generates thousands of random masks and then employ them to mask the input. Then a linear combination of random masks with the corresponding prediction score of the masked images is computed as the final saliency map. Instead of generating random masks, Score-CAM~\cite{DBLP:conf/cvpr/WangWDYZDMH20} adopts feature maps of the target layer (the target layer generally contains thousands of feature maps) as initial masks and employ them to computing saliency map. Unlike RISE and Score-CAM, XRAI~\cite{DBLP:conf/iccv/KapishnikovBVT19} first over-segmented the input image, and then iteratively test the importance of each region, coalescing smaller regions into larger segments based on attribution scores. Region-based approaches usually generate better human interpretable visualizations but are less efficient since they requires plenty of quires to the neural network~\cite{DBLP:conf/aaai/ZhangRY21}. Our Group-CAM can be seen as a region-based method while is much faster than RISE~\cite{DBLP:conf/bmvc/PetsiukDS18}, Score-CAM~\cite{DBLP:conf/cvpr/WangWDYZDMH20} and XRAI~\cite{DBLP:conf/iccv/KapishnikovBVT19}.

\textbf{Activation-based Saliency Methods.}
These approaches combine activations (generally the combination of back-propagation gradients and feature maps) of a selected convolutional layer to form an explanation~\cite{DBLP:conf/cvpr/ZhouKLOT16,DBLP:conf/iccv/SelvarajuCDVPB17,DBLP:conf/wacv/ChattopadhyaySH18}. CAM~\cite{DBLP:conf/cvpr/ZhouKLOT16} and Grad-CAM~\cite{DBLP:conf/iccv/SelvarajuCDVPB17} adopt a linear combination of activations to form a heatmap with fine-grained details. Grad-CAM++~\cite{DBLP:conf/wacv/ChattopadhyaySH18} extends Grad-CAM and uses a weighted combination of the positive partial derivatives of the target layers’ feature maps with respect to a specific class score as weights to generate a visual explanation for the corresponding class label.
Activation-based methods are in general faster than region-based approaches since they only require a single or constant number of queries to the model~\cite{DBLP:conf/iccv/KapishnikovBVT19}. However, results of activation-based methods may capture too much meaningless information since the feature maps are not necessarily related to the target category~\cite{DBLP:conf/aaai/ZhangRY21}. In this paper, we draw on the idea of Grad-CAM~\cite{DBLP:conf/iccv/SelvarajuCDVPB17} to generate initial-mask of Group-CAM.


\section{Group-CAM}
\label{sec:groupcam}
In this section, we first describe the Group-CAM algorithm, then explain the motivation behind it. The high-level steps are shown in Algorithm~\ref{alg:alg-1}.

\begin{algorithm}[htb]
	\caption{Group-CAM Algorithm}
	\label{alg:alg-1} 
	\KwIn{Image $I_0$, Model $\mathcal{F}$, Class $c$, number of groups $G$, Gaussian blur parameters: $ksize$, $sigma$.}
	\KwOut{Saliency map $\mathcal{L}_{Group-CAM}^{c}$}
	Initialization: Initial $\mathcal{L}_{Group-CAM}^{c}\gets 0$, Baseline Input $\tilde{I_0} = guassian\_blur2d(I_0, ksize, sigma)$, $\ell \gets 0$\;
	Get target layer feat maps $A$, importance weights $w^{c}$\;
	$K \gets$ the number of channels of $A$\;
	$g \gets K/G$ number of feat maps in each group\;
	\While{$\ell < G$}{
		Generating $M_{\ell} = ReLU(\sum_{k={\ell}\times g}^{({\ell}+1)\times g-1}(w_{k}^{c}A^k))$\;
		Initial mask $M'_{\ell} \gets$ de-nosie, normalize and bilinear interpolate upsample activation map $M_{\ell}$\;
		Perturbated image $I'_\ell = I_0 \odot M'_{\ell} + \tilde{I_0} \odot (1 - M'_{\ell})$\;
		Compute confidence gain $\alpha_{\ell}^{c} = \mathcal{F}_c(I'_\ell) - \mathcal{F}_c(\tilde{I_0})$\;
		$\mathcal{L}_{Group-CAM}^{c} \gets \mathcal{L}_{Group-CAM}^{c} + \alpha_{\ell}^{c} M'_{\ell}$ \;
		$\ell \leftarrow \ell+1$\;
	}
	\Return ${\rm} ReLU(\mathcal{L}_{Group-CAM}^{c})$
\end{algorithm}

\subsection{Initial Masks}
Let $I_0 \in \mathbb{R}^{3\times H \times W}$ be an input image, $\mathcal{F}$ be a deep neural network which predicts a score $\mathcal{F}_{c}(I_0)$ on class $c$ with input $I_0$. In order to obtain the class-discriminative initial group masks of the target convolutional layer, we first compute the gradient of $\mathcal{F}_{c}(I_0)$ with respect to $k^{th}$ feature map $A^{k}$. Then these gradients are global average pooled over the height and width dimensions (indexed by $i, j$ respectively) to obtain the neuron importance weights

\begin{equation}
	\label{eq:eq1}
	w_{k}^{c} = \dfrac{1}{Z}\sum_{i}\sum_{j}\dfrac{\partial \mathcal{F}_c(I_0)}{\partial A_{ij}^{k}(I_0)}
\end{equation}
where $Z$ is the number of pixels in feature map $A^{k}$.

Assume $K$ is the number of channels of the target layer feature maps, we first split all the feature maps and neuron importance weights into $G$ groups. Then the initial mask in each group is calculated by

\begin{equation}
	\label{eq:eq2}
	M_{\ell} = ReLU(\sum_{k={\ell}\times g}^{({\ell}+1)\times g-1}(w_{k}^{c}A^k))
\end{equation}
where $\ell \in \{0, 1, \cdots, G-1\}$, $g=K/G$ is the number of feature maps in each group.

$M_{\ell}$ is the combination of feature maps and gradients, which means $M_{\ell}$ can be noisy visually since the gradients for a DNN tend to vanish due to the flat zero-gradient region in ReLU~\cite{DBLP:conf/aaai/ZhangRY21}. Therefore, it is not suitable to directly apply $M_{\ell}$ as the initial mask.

To remedy this issue, we utilize a de-noising function to filter pixels in $M_\ell$ less than $p(M_\ell, \theta)$, where $p(M_\ell, \theta)$ is a function which compute the $\theta^{th}$ percentile of $M_\ell$.

Formally, for a scalar $m_{ij}$ in $M_{\ell}$, the de-noising function can be represented as
\begin{equation}
	\label{eq:eq3}
	\phi(m_{ij}, \theta) = \begin{cases}
		m_{ij}, & \text{if } m_{ij} > p(M_\ell, \theta); \\
		0, & \text{otherwise}.
	\end{cases}
\end{equation}

Instead of setting all pixels to binary values, it is better to generate smoother mask for an activation map. Specifically, we scale the raw values of $M_\ell$ into $[0, 1]$ by utilizing Min-Max normalization, 
\begin{equation}
	\label{eq:eq4}
	M'_\ell = \frac{M_\ell - min(M_\ell)}{max(M_\ell) - min(M_\ell)}
\end{equation}
Then, $M'_\ell$ is upsampled with bilinear interpolate to the same resolution of $I_0$ to mask the input. 

\subsection{Saliency Map Generation}
It has been widely acknowledged that if the saliency method is in fact identifying pixels significant to the model's prediction, this should be reflected in the model's output for the reconstructed image~\cite{DBLP:conf/iccv/KapishnikovBVT19}. However, merely masking the image pixels out of the region of interest causes unintended effects due to the sharp boundary between the masked and salient region. Consequently, it is crucial to minimize such adversarial effects when testing for the importance of a feature subset \cite{DBLP:conf/nips/DabkowskiG17}.

To address this issue, we start with a masked version of the input, replace the unreserved regions (pixels with 0 values) with the blurred information, and then performing classification on this image to measure the importance of the initial masks. The blurred images can be computed by
\begin{equation}
	\label{eq:eq5}
	I'_\ell = I_0 \odot M'_\ell + \tilde{I_0} \odot (1-M'_\ell)
\end{equation}
where $\tilde{I_0} = guassian\_blur2d(I_0, ksize, sigma)$ is a baseline image with the same shape as $I_0$ and have a lower confidence of class $c$. In this paper, we set $ksize=51$ and $sigma=50$, following \cite{DBLP:conf/aaai/ZhangRY21}.

The contribution $\alpha_{\ell}^{c}$ of the reserved regions $I_0 \odot M'_\ell$ can then be computed as
\begin{equation}
	\label{eq:eq6}
	\alpha_{\ell}^{c} = \mathcal{F}_c(I'_\ell) - \mathcal{F}_c(\tilde{I_0})
\end{equation}

Similar to RISE~\cite{DBLP:conf/bmvc/PetsiukDS18}, the final saliency map is a linear combination of the initial masks with weights $\alpha_{\ell}^{c}$, that is
\begin{equation}
	\label{eq:eq7}
	\mathcal{L}_{Group-CAM}^{c} = ReLU(\sum_{\ell} \alpha_{\ell}^{c} M'_\ell)
\end{equation}

%

\begin{figure*}[htb]
	\centering
	\includegraphics[width=1.0\linewidth]{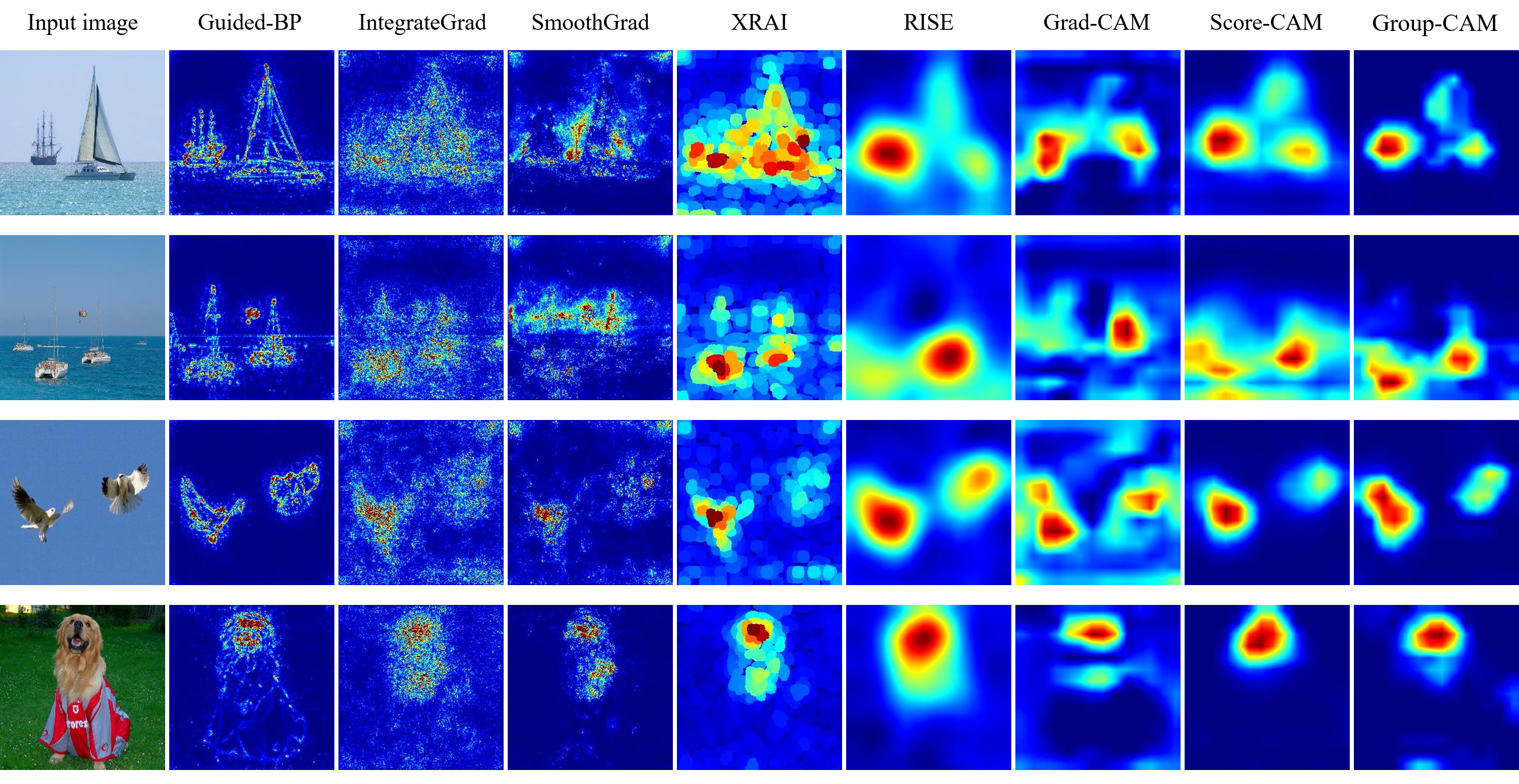}
	\caption{Visualization results of SOTA saliency methods. Results show that saliency maps of Group-CAM are more compelling than region-base methods and activation-based methods, and contain less noise than gradient-based methods.}
	\label{fig:fig3}
\end{figure*}

\section{Experiments}
In this section, we firstly utilize ablation studies to investigate the effect of group $G$ and filtering threshold $\theta$. Then we apply a sanity check to test whether Group-CAM is sensitive to model parameters. Finally, we compare the proposed Group-CAM with other popular saliency methods to evaluate its performance.

\subsection{Experimental Setup}
Experiments in this section are conducted on the commonly-used computer vision datasets, including the validation split of ImageNet-1k~\cite{DBLP:journals/ijcv/RussakovskyDSKS15}(containing 50k images) and the val2017 split of MS COCO2017~\cite{DBLP:conf/eccv/LinMBHPRDZ14} (containing 5k images). For both datasets, all images are resized to $3\times224\times224$, and then transformed to tensors and normalized to the range [0, 1]. No further pre-processing is performed. We report the insertion and deletion test results using the pre-trained torchvision model\footnote{https://github.com/pytorch/vision/tree/master/torchvision} VGG19~\cite{DBLP:journals/corr/SimonyanZ14a} as the base classifier and other results are on the pre-trained ResNet-50~\cite{DBLP:conf/cvpr/HeZRS16}. Unless explicitly stated, the number of groups $G$ adopted in Group-CAM is 32, and the threshold $\theta$ in Eq.~\ref{eq:eq3} is set as 70. For a fair comparison, all saliency maps are upsampled with bilinear interpolate to $224 \times 224$.

\subsection{Class Discriminative Visualization}
We qualitatively compare the saliency maps produced by recently SOTA methods, including gradient-based methods (Guided Backpropagation~\cite{DBLP:journals/corr/SpringenbergDBR14}, IntegrateGrad~\cite{DBLP:conf/icml/SundararajanTY17}, SmoothGrad~\cite{DBLP:journals/corr/SmilkovTKVW17}), region-based methods(RISE~\cite{DBLP:conf/bmvc/PetsiukDS18}, XRAI~\cite{DBLP:conf/iccv/KapishnikovBVT19}), and activation-based methods (Grad-CAM~\cite{DBLP:conf/iccv/SelvarajuCDVPB17}, Score-CAM~\cite{DBLP:conf/cvpr/WangWDYZDMH20}) to validate the effectiveness of Group-CAM.

As shown in Figure~\ref{fig:fig3}, results in Group-CAM, random noises are much less than that in region-base methods and activation-base methods. In addition, Group-CAM generates smoother saliency maps comparing with gradient-based methods.

\begin{figure}[h]
	\centering
	\includegraphics[width=0.98\linewidth]{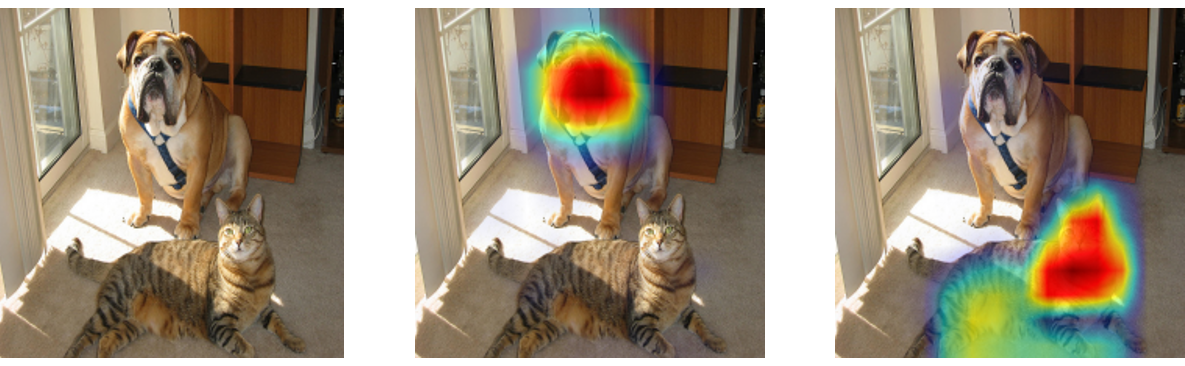}
	\caption{Class discriminative results. The middle image is generated w.r.t `bull mastiff', and the right one is generated w.r.t `tiger cat'.}
	\label{fig:fig5}
\end{figure}

We further conduct experiments to test whether that Group-CAM can distinguish different classes. As shown in Figure~\ref{fig:fig5}, the VGG19 classifies the input as ``bull mastiff" with 46.06\% confidence and `tiger cat' with 0.39\% confidence. Group-CAM correctly gives the explanation locations for both of two categories, even though the classification score of the latter is much lower than that of the former. It is reasonable to indicate that Group-CAM can distinguish different categories. 

\begin{figure*}[htb]
	\centering
	\includegraphics[width=1.0\linewidth]{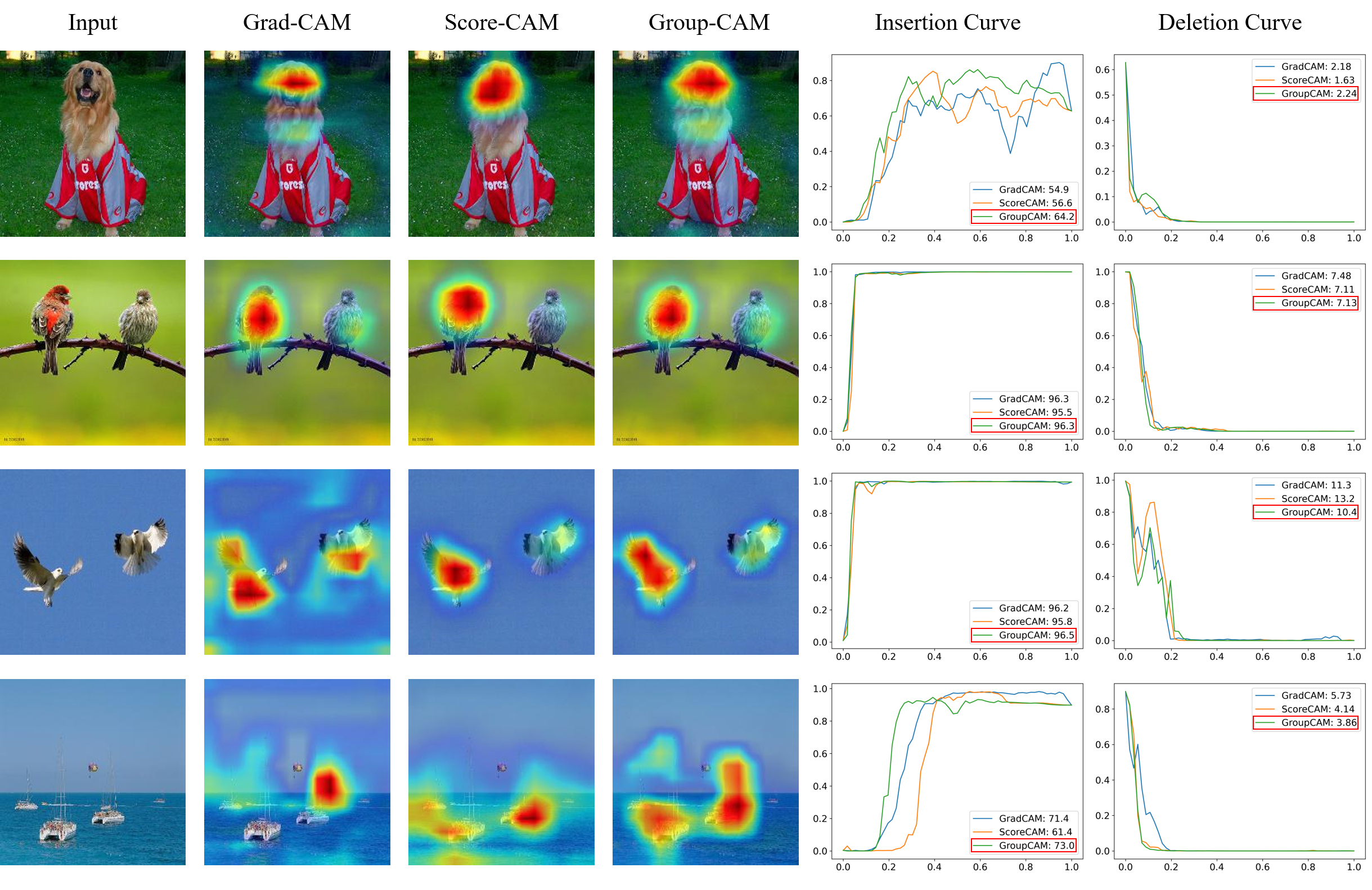}
	\caption{Grad-CAM, Score-CAM and Group-CAM generated saliency maps for representative images in terms of deletion and insertion curves. In the insertion curve, a better explanation is expected that the prediction score to increase quickly, while in the deletion curve, it is expected the classification confidence to drop faster.}
	\label{fig:fig2}
\end{figure*}

\begin{table*}[htb]
		\begin{center}
			\begin{tabular}{ccccccc}
				\toprule[1.2pt]
				AUC & Grad-CAM & Grad-CAM++ & RISE & XRAI & Score-CAM & \textbf{Group-CAM} \\
				\midrule[1.2pt]
				Insertion & 53.5 & 50.0 & 54.0  & 43.1  & 55.1  & \textbf{56.8}  \\
				\midrule
				Deletion & 13.3 & 14.8 & 11.7  & 15.8  & \textbf{11.5}  & 12.3  \\
				\midrule
				Over-all & 40.2 & 35.2 & 42.3 & 27.3 & 43.6 & \textbf{44.5} \\ 
				\bottomrule[1.2pt]
			\end{tabular}
		\end{center}
	\caption{Comparative evaluation in terms of deletion (\textbf{lower AUC is better}) and insertion (\textbf{higher AUC is better}) AUC on ImageNet-1k validation split. The over-all score (\textbf{higher AUC is better}) shows that Group-CAM outperform other related methods significantly. The best records are marked in \textbf{bold}.}
\label{table-2}
\end{table*}

\subsection{Deletion and Insertion}
We follow ~\cite{DBLP:conf/bmvc/PetsiukDS18} to conduct deletion and insertion tests to evaluate different saliency approaches. The intuition behind the deletion metric is that the removal of pixels/regions most relevant to a class will cause the classification score to drop significantly~\cite{DBLP:conf/bmvc/PetsiukDS18,DBLP:journals/tnn/SamekBMLM17}. Insertion metric, on the other hand, starts with a blurred image and gradually re-introduces content, which produces more realistic images and has the additional advantage of mitigating the impact of adversarial attack examples~\cite{DBLP:conf/bmvc/PetsiukDS18}.
In detail, for the deletion test, we gradually replace 3.6\% (i.e., $224\times8$) pixels in the original image with a highly blurred version each time according to the values of the saliency map until no pixels left. Contrary to the deletion test, the insertion test replaces 3.6\% pixels of the blurred image with the original one until the image is well recovered. We calculate the AUC of the classification score after Softmax as a quantitative indicator. 
Besides, we provide the $over-all$ score to comprehensively evaluate the deletion and insertion results, which can be calculated by $AUC(insertion) - AUC(deletion)$. Examples are shown in Figure~\ref{fig:fig2}. The average results over all the 50k images is reported in Table~\ref{table-2}.

As illustrated in Table~\ref{table-2}, the proposed Group-CAM outperforms other related approaches in terms of insertion and over-all AUC. Moreover, Group-CAM also exceeds other methods in terms of deletion AUC except for XRAI. 

\begin{figure*}[htb]
	\centering
	\includegraphics[width=0.98\linewidth]{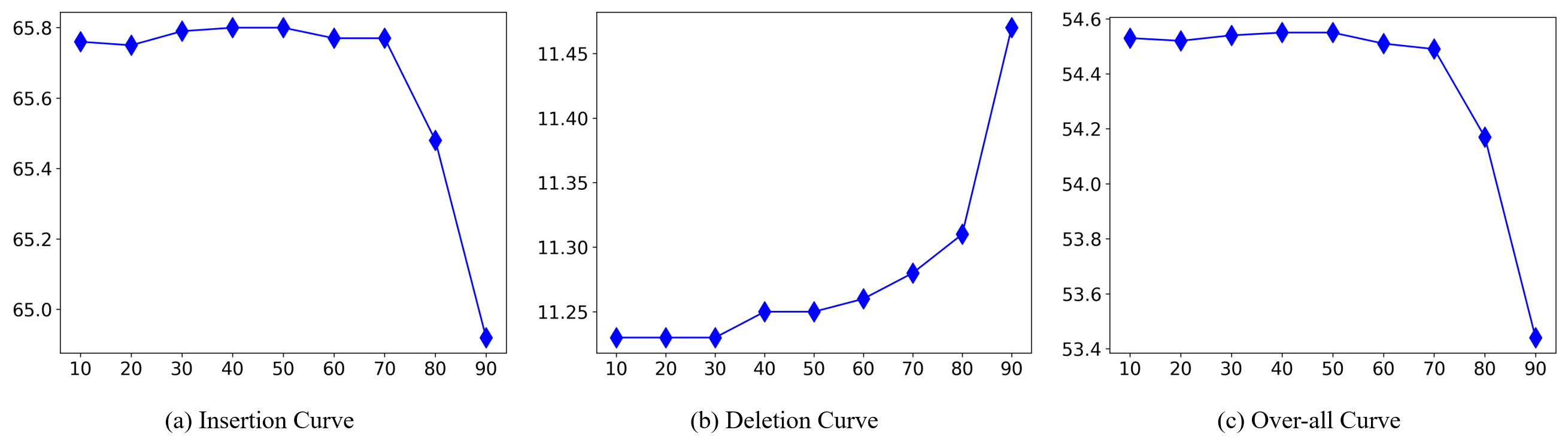}
	\caption{Ablation studies of filter threshold $\theta$ with $G=32$ in terms of deletion (\textbf{lower AUC is better}), insertion (\textbf{higher AUC is better}) curve and the over-all scores (\textbf{higher AUC is better}) on ImageNet-1k validation split(on the first 5k images).}
	\label{fig:fig6}
\end{figure*}

\textbf{Ablation Studies.}
We report the ablation studies results of Group-CAM on the first 5k images on ImageNet-1k, to thoroughly investigate the influence of filter threshold $\theta$ and group $G$. Results are shown in Figure~\ref{fig:fig6} and Table~\ref{table-1}.
\begin{table}[h]
	\setlength{\tabcolsep}{12.0pt}{
		\begin{center}
			\begin{tabular}{c c c c}
				\toprule[1.2pt]
				Groups & Insertion & Deletion & Over-all \\
				\midrule[1.2pt]
				1 & 61.72  & \textbf{11.21}  & 50.51  \\
				\midrule 
				4 & 64.27  & 11.21  & 53.07  \\
				\midrule
				8 & 64.94  & 11.29  & 53.65  \\
				\midrule
				16 & 65.38  & 11.34  & 54.04  \\
				\midrule
				32 & 65.48  & 11.31  & 54.17  \\ 
				\midrule
				64 & 65.77  & 11.31  & 54.46  \\
				\midrule
				128 & 65.81  & 11.29  & 54.52  \\
				\midrule
				256 & \textbf{65.84}  & 11.28  & \textbf{54.56}  \\ 
				\bottomrule[1.2pt]
			\end{tabular}
		\end{center}
	}
	\caption{Ablation studies of Group $G$ with filter threshold $\theta=70$ in terms of deletion, insertion, and over-all scores on ImageNet-1k validation split (on the first 5k images). The best records are marked in \textbf{bold}.}
	\label{table-1}
\end{table}

From Figure~\ref{fig:fig6}, we can see, threshold $\theta$ has a significant effect on the performance of Group-CAM (fluctuating over 1.1\% in terms of over-all score). Specifically, when $\theta$ is small, the over-all score keeps stable with an increase of $\theta$. Then, as $\theta$ increases, the over-all score drops quickly when $\theta>70$. Here, to make a trade-off between insertion and deletion results, we set $\theta=70$ as default.

Besides, in Table~\ref{table-1} we can see, the over-all score increase with the increase of $G$. However, as introduced in Algorithm~\ref{alg:alg-1}, larger $G$ means more computer costing. To make a trade-off, we set $G=32$ as the default group of Group-CAM.

\begin{table}[htb]
	\setlength{\tabcolsep}{12.0pt}{
		\begin{center}
			\begin{tabular}{cc}
				\toprule[1.2pt]
				Methods & Running Time \\ 
				\midrule[1.2pt]
				RISE & 38.23 \\
				\midrule
				XRAI & 42.17 \\
				\midrule
				\textbf{Grad-CAM} & \textbf{0.03} \\
				\midrule
				Score-CAM & 2.46 \\
				\midrule
				Group-CAM (ours) & \textcolor[rgb]{ 0,  0,  1}{0.09} \\
				\bottomrule[1.2pt]
			\end{tabular}
	\end{center}}
\caption{Comparative evaluation in terms of running time (seconds, averaged on 5k images) on ImageNet-1k. The best and second best records are marked in \textbf{bold} and \textcolor[rgb]{ 0,  0,  1}{blue}, respectively.}
\label{table-3}
\end{table}

\textbf{Running Time.}
In Table~\ref{table-3}, we summarize the average running time for RISE~\cite{DBLP:conf/bmvc/PetsiukDS18}, XRAI~\cite{DBLP:conf/iccv/KapishnikovBVT19}, Grad-CAM~\cite{DBLP:conf/iccv/SelvarajuCDVPB17}, Score-CAM~\cite{DBLP:conf/cvpr/WangWDYZDMH20} and the proposed Group-CAM on one NVIDIA 2080Ti GPU. 
As shown in Table~\ref{table-3}, the averaging running time for Grad-CAM and Group-CAM are both less than 1 second, which achieve best results among all the approaches. Combined with Table~\ref{table-2} and Table~\ref{table-3}, we observe that although Group-CAM runs slower than Grad-CAM, it achieves much better performance.

\subsection{Localization Evaluation}
In this part, we adopt pointing game~\cite{DBLP:journals/ijcv/ZhangBLBSS18} on the val2017 of MS COCO2017 to measure the quality of the generated saliency map through localization ability. We apply the same pre-trained ResNet-50 from \cite{DBLP:conf/bmvc/PetsiukDS18}. The localization accuracy is then calculated as $Acc=\dfrac{\#Hits}{\#Hits + \#Misses}$ for each object category (if the most salient pixel lies inside the annotated bounding boxes of an object, it is counted as a hit). The overall performance is measured by the mean accuracy across different categories.

From Table~\ref{table-4}, we observe that Group-CAM beats all the other compared approaches. Specifically, Group-CAM outperforms the base Grad-CAM with 0.8\% in terms of mean accuracy.

\begin{table}[htb]
	\setlength{\tabcolsep}{16.0pt}{
		\begin{center}
			\begin{tabular}{cc}
				\toprule[1.2pt]
				Methods & Mean Accuracy \\
				\midrule[1.2pt]
				Grad-CAM & 56.7  \\
				\midrule
				Grad-CAM++ & 57.2  \\
				\midrule
				RISE & 54.3  \\
				\midrule
				XRAI & 55.1  \\
				\midrule
				Score-CAM & 51.0  \\
				\midrule
				\textbf{Group-CAM (ours)} & \textbf{57.5}  \\
				\bottomrule[1.2pt]
			\end{tabular}
	\end{center}}
\caption{Pointing Game on COCO val2017 split. Results show that the proposed Group-CAM performs consistently better than other related methods.}
\label{table-4}
\end{table}

\subsection{Sanity Check}
\begin{figure}[htb]
	\centering
	\includegraphics[width=1.0\linewidth]{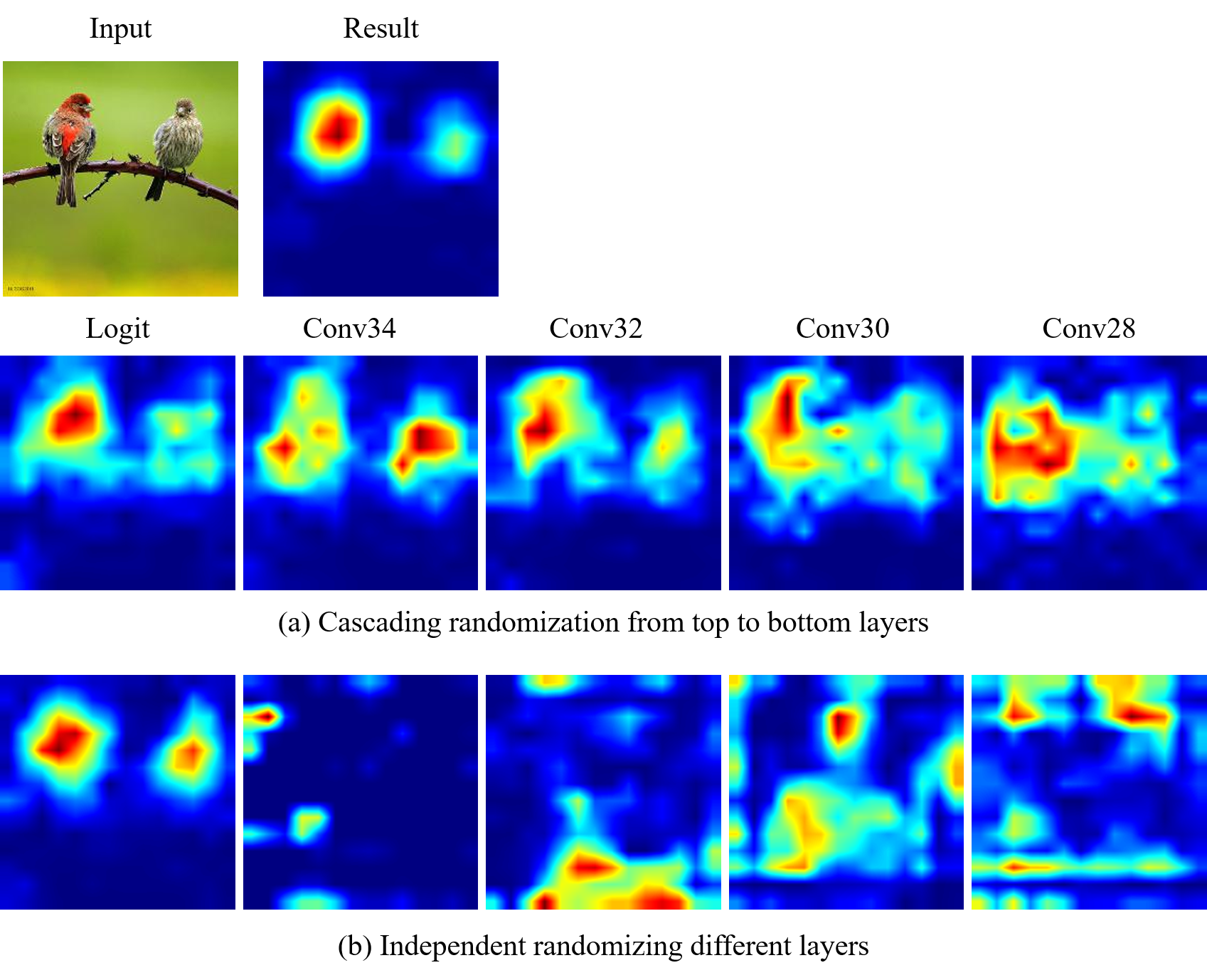}
	\caption{Sanity check results by cascade randomization and independent randomization. Results show that Group-CAM is sensitive to classification model parameters and can reflect the quality of the network.}
	\label{fig:fig4}
\end{figure}

Finally, we utilize sanity check \cite{DBLP:conf/nips/AdebayoGMGHK18} to check whether the results of Group-CAM can be considered completely reliable explanations for a trained model's behavior. Specifically, we employ both cascade randomization and independent randomization, to compare the output of Group-CAM on a pre-trained VGG19. As shown in Figure~\ref{fig:fig4}, The Group-CAM is sensitive to classification model parameters and can produce valid results.

\section{Fine-tuning Classification Methods}
Finally, we extend the application of Group-CAM and apply it as an effective data augment strategy to fine-tune/train the classification models. We argue that a saliency method that is suitable to fine-tune the networks should have the following two characteristics: (1) the saliency method should be efficient, which can produce saliency maps in limited times; (2) the generated saliency maps should be related to the object targets. Our Group-CAM can produce appealing target-related saliency maps in 0.09 seconds per image with $G=32$, which means Group-CAM is suitable to fine-tune the networks. 
 
To make Group-CAM more efficient, we remove the importance weights $w^{c}$ and de-noise procedures. Although this will slightly impair the performance of Group-CAM,  back-propagation is no longer needed, which can greatly save the saliency maps generating time. 

The fine-tuning process is defined as follows:

(1) generate saliency map $M$ for $I_0$ with $G=16$ and the ground-truth target class $c$;

(2) binarize $M$ with threshold $\theta$, where $\theta$ is the mean value of $M$.

(3) apply Eq.~\ref{eq:eq5} to get the blurred input $\tilde{I_0}$.

(4) adopt $\tilde{I_0}$ to fine-tune the classification model.

Since $\tilde{I_0}$ are generated during the training process, which means that when the performance of the classification model is improved, Group-CAM will generate a better $\tilde{I_0}$, which in turn will promote the performance of the classification model.

\begin{figure}[htb]
	\centering
	\includegraphics[width=0.80\linewidth]{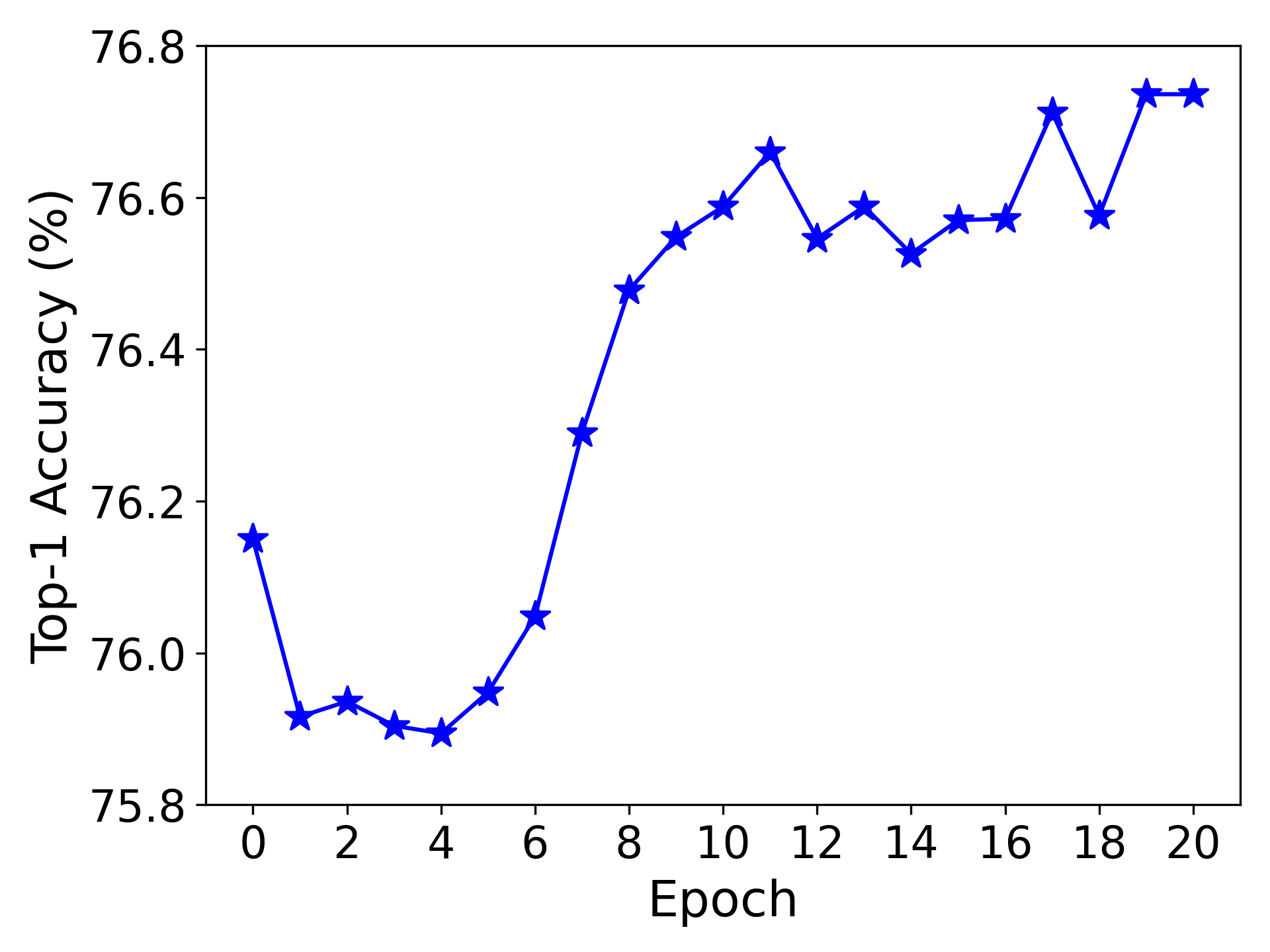}
	\caption{Fine-tuning ResNet-50 with Group-CAM. Results show that Group-CAM can improve the classification model's performance by a significant margin.}
	\label{fig:fig7}
\end{figure}

Here, we report the results on the ImageNet-1k validation split of fine-tuning ResNet-50. Specifically, we trained the pre-trained ResNet-50 by SGD with weight decay 1e-4, momentum 0.9, and mini-batch size 256 (using 8 GPUs with 32 images per GPU) for 20 epochs, starting from the initial learning rate of 1e-3 and decreasing it by a factor of 10 every 15 epochs. For the testing on the validation set, the shorter side of an input image is first resized to 256, and a center crop of $224 \times 224$ is used for evaluation. 

As shown in Figure~\ref{fig:fig7}, fine-tune with Group-CAM can contribute to 0.59\% (76.74\% vs. 76.15\%) improvement in terms of Top-1 accuracy. 

\begin{figure}[htb]
	\centering
	\includegraphics[width=1.0\linewidth]{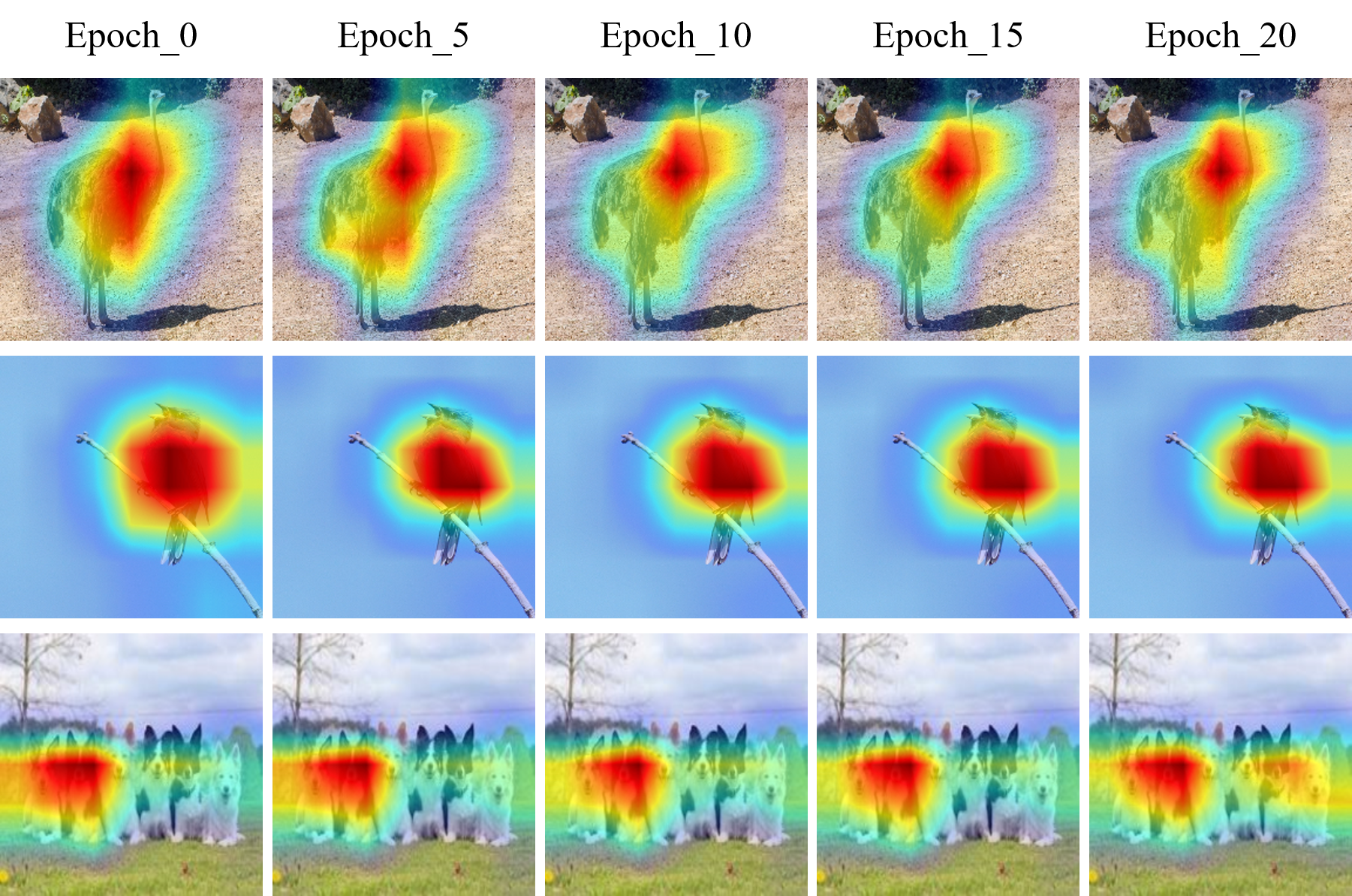}
	\caption{Visualization results of fine-tuning ResNet-50 with Group-CAM. The first image (Epoch\_0) is generated the original pre-trained ResNet-50. The right four images (i.e., Epoch\_5, Epoch\_10, Epoch\_15 and Epoch\_20) are generated by the fine-tuning ResNet-50.}
	\label{fig:fig8}
\end{figure}

Here, we visualize the saliency maps generated by the fine-tuned ResNet-50 in Figure~\ref{fig:fig8}. As illustrated in Figure~\ref{fig:fig8}, as the performance of ResNet-50 improves, the noise of the saliency maps generated by Group-CAM decreases and focuses more on the important regions. Since the noise can reflect the performance to some degree, we can also treat it as a hint to determine whether a model has converged. That is, if the saliency maps generated by Group-CAM do not change, the model may have converged.

\section{Conclusion}
In this paper, we proposed Group-CAM, which adopts the grouped sum of gradient and feature map combinations as initial masks. These initial masks are adopted to preserve a subset of input pixels, and then these pixels are fed into the network to calculate the confidence scores, which reflects the importance of the masked images. The final saliency map of Group-CAM is computed as a weighted sum of the initial masks, where the weights are confidence scores produced by the masked inputs. The proposed Group-CAM is efficient yet effective and can be applied as a data augment trick to fine-tune/train classification models.  
Experimental results on ImageNet-1k and COCO2017 demonstrate that Group-CAM achieves better visual performance than the current state-of-the-art explanation approaches.


{\small
\bibliographystyle{ieee_fullname}
\bibliography{ref}
}

\end{document}